\documentclass[11pt, a4paper, copyright, gdm]{google}

\usepackage{url}
\usepackage{tikz}
\usepackage{csquotes}
\usetikzlibrary{positioning}

\usepackage[
    backend=biber,
    style=numeric-comp, %
    sorting=none,       %
    giveninits=true, 
    uniquename=init,
    natbib=true,
    maxbibnames=99,   %
    maxcitenames=2,   %
    mincitenames=1    %
]{biblatex}

\DeclareSourcemap{
  \maps[datatype=bibtex]{
    \map{
      \step[fieldset=primaryclass, null]
    }
  }
}

\DeclareNameAlias{sortname}{given-family}

\RequirePackage[colorlinks=true, allcolors=blue]{hyperref}
\usepackage[capitalize]{cleveref}

\DeclareNameAlias{sortname}{given-family}

\newboolean{USECARTOON}
\setboolean{USECARTOON}{false} %

\newcommand{\usescreenshot}[2]{%
  \ifthenelse{\boolean{USECARTOON}}{%
    \includegraphics[#1]{assets/screenshot-#2-CARTOON.jpeg}%
  }{%
    \includegraphics[#1]{assets/screenshot-#2.png}%
  }%
}

\uselogo{} 

\title{{AI co-mathematician: Accelerating mathematicians with agentic AI}}

\correspondingauthor{dhhzheng@google.com, pushmeet@google.com}

\reportnumber{249243} %

\renewcommand{\today}{2026-05-07}

\author[1]{Daniel Zheng}
\author[1]{Ingrid von Glehn}
\author[1]{Yori Zwols}
\author[1]{Iuliya Beloshapka}
\author[1]{Lars Buesing}
\author[1]{Daniel M.~Roy}
\author[1]{Martin Wattenberg}
\author[1]{Bogdan Georgiev}
\author[1]{Tatiana Schmidt}
\author[1]{Andrew Cowie}
\author[1]{Fernanda Viegas}
\author[1]{Dimitri Kanevsky}
\author[2]{Vineet Kahlon}
\author[1]{Hartmut Maennel}
\author[1]{Sophia Alj}
\author[1]{George Holland}
\author[1]{Alex Davies}
\author[1]{Pushmeet Kohli}

\affil[1]{\thepa{}{}}
\affil[2]{Google}

\newcommand{\user}[1]{\emph{#1}}

\begin{abstract}
We introduce the AI co-mathematician, a workbench for mathematicians to interactively leverage AI agents to pursue open-ended research. The AI co-mathematician is optimized to provide holistic support for the exploratory and iterative reality of mathematical workflows, including ideation, literature search, computational exploration, theorem proving and theory building. By providing an asynchronous, stateful workspace that manages uncertainty, refines user intent, tracks failed hypotheses, and outputs native mathematical artifacts, the system mirrors human collaborative workflows. In early tests, the AI co-mathematician helped researchers solve open problems, identify new research directions, and uncover overlooked literature references. Besides demonstrating a highly interactive paradigm for AI-assisted mathematical discovery, the AI co-mathematician also achieves state of the art results on hard problem-solving benchmarks, including scoring 48\% on FrontierMath Tier 4, a new high score among all AI systems evaluated. 
\end{abstract}

\begin{document}

\maketitle

\section{Introduction}

Mathematics research is a multi-dimensional, complex, highly iterative process.
While the publication record is built almost entirely on polished, rigorous proofs, a mathematician's day-to-day work has long been understood to involve activities traditionally concealed from public view \citep{polya1954, lakatos1976, epstein1992}.
Beneath the final formalism lies a deeply exploratory process, where initial intuitions are tested, counter-examples are discovered, and both core definitions and proofs are subjected to continuous cycles of refutation and revision.

Recent years have seen an explosion of capabilities in AI-for-mathematics,
rapidly expanding the field across several distinct dimensions.
In the realm of autonomous mathematical reasoning, building on early systems like Minerva~\citep{lewkowycz2022minerva} and the vast body of subsequent work
\citep{taylor2022galactica,azerbayev2023llemma,yue2023mammoth,shao2024deepseekmath,mistral2024mathstral,yang2024qwen25math,lu2024aiscientist,deepthink2025,he2025skywork,yamada2025aiscientistv2,liu2025deepevolve,zimmer2026agentic}, the field has now advanced to breakthroughs like the autonomous research system Aletheia~\citep{feng2026}.
In the realm of exploratory search, AlphaEvolve \citep{romeraparedes2024, novikov2025} and systems it has inspired \citep{cemri2026adaevolve,sharma2025openevolve,lange2025shinkaevolve} allow researchers to uncover novel algorithms and structures through continuous, steerable evolutionary runs.
In the realm of formalized mathematics, AlphaProof \citep{hubert2025}, related systems \citep{song2024leancopilot, wang2024legoprover, ren2025deepseekproverv2, lin2025goedelproverv2, li2025lips, chen2025seedprover, wang2025kimina, hariharan2026milestoneformalizationspherepacking, li2024hunyuanprover, jana2025proofbridge, chen2026parity, liu2026numina, deltredici2025axprover},
and interactive environments such as Aristotle \citep{harmonic2025aristotle} have deeply integrated reinforcement learning and language models into verified mathematics and open-source proof assistants.
Meanwhile, 
the inference-scaling models currently powering commercial chat interfaces have already brought immense problem-solving power directly into the hands of mathematicians  \citep{woodruff2026accelerating, bubeck2025early, alexeev2026short1, alexeev2026short2, gottweis2025towards}.

Yet first-hand experience with these systems reveals that a critical dimension of mathematical research remains under-supported: the orchestration of these capabilities into a long-term, stateful, collaborative workflow.
Day-to-day mathematics is rarely a sequence of isolated queries or computer-assisted proofs; it involves managing uncertainty, synthesizing disparate literature, drafting and revising intermediate artifacts, and tracking complex, branching hypotheses over days or weeks. 

Because standard chat interfaces are inherently transient and specialized engines lack broader context, researchers must act as the manual connective tissue between conversational brainstorming, formal provers, and computational scripts.
While software developers increasingly rely on AI-powered coding environments to provide this exact kind of orchestration, those tools are fundamentally optimized for the lifecycle of code, not the unique abstractions, proofs, and artifacts of mathematics.
To truly accelerate discovery, we believe the next phase of AI-for-mathematics must focus on this missing dimension of orchestration, natively supporting the full, messy reality of mathematical activities.

To understand how to build a native environment for these activities, we can look deeper into the field of software engineering.
Recent advancements (for example, Google Antigravity \citep{google2025antigravity}, Claude Code \citep{anthropic2025claudecode}, and OpenAI Codex \citep{openai2025codex}) have demonstrated the transformative potential of efficient interactive collaboration between humans and agentic AI systems.
One of the reasons behind the success of these systems is that pre-existing software engineering practices embody the paradigms necessary to capture and accelerate iterative exploration.
Informal software specifications such as design documents allow agents to work autonomously for long periods of time whilst remaining on a predefined path, and continuous testing pipelines provide an automated flow for verification, while version control seamlessly tracks and maintains the evolving state of a project in a manner familiar to human professionals.
In contrast, almost none of the analogous activities in the everyday work of a mathematician are routinely automated.

\begin{quote}\em
    To establish this missing agentic flow for mathematics, we introduce the AI co-mathematician, a workbench for mathematicians to interactively leverage AI agents to pursue open-ended research, based on the latest Gemini language models.
\end{quote}
The AI co-mathematician provides a stateful, interactive workspace where a project coordinator agent delegates complex tasks across parallel workstreams, allowing the user to direct and interact with an evolving research process, rather than waiting for end-to-end autonomous execution.
As with other agentic systems, the AI co-mathematician is a harness designed to be used with standard, commercially available language models, and does not rely on any custom model behavior or training. 
This architectural direction parallels recent momentum in empirical domains to construct dedicated, multi-agent workbenches optimized for open-ended hypothesis generation and scientific data analysis \citep{gottweis2025towards,mitchener2025kosmos}.

Importantly, the AI co-mathematician is designed to complement existing frontier approaches rather than replace them.
By establishing this stateful architecture, we pave the way for powerful underlying engines---whether autonomous reasoners like AlphaProof and Aletheia, or evolutionary iterators like AlphaEvolve---to be deployed dynamically within the human researcher's interactive loop.
While the AI co-mathematician itself is currently subject to a limited initial release, our goal is to develop future products that grant much broader access to this interactive paradigm.

In the remainder of this paper, we detail the philosophy and architecture behind the AI co-mathematician, and share early results.
Section 2 outlines our core design principles for interactive, AI-assisted mathematics.
Section 3 grounds these principles in practice through a concrete walkthrough, highlighting the programmatic constraints and adversarial review loops that prevent the system from taking the easy path on intractable problems, and the parallel workstreams which enable multi-faceted research.
Section 4 discusses evaluation strategies for interactive mathematical agents such as the AI co-mathematician.
Section 5 presents early qualitative results, demonstrating how practicing mathematicians have used the system to steer open-ended research and generate verifiable insights.
Section 6 evaluates the system's performance on static problem-solving benchmarks, providing a baseline measurement of its underlying capabilities.
Section 7 describes challenges and limitations we encountered in developing the AI co-mathematician.
Finally, we conclude by outlining our vision for the next era of AI-assisted mathematics.

\section{Design Principles for AI-Assisted Mathematics}

The overriding goal of our design is to support human mathematicians in their own work.
As \citet{Thurston1994} famously argued, mathematics is fundamentally a social enterprise aimed at advancing human understanding, rather than merely a machine for generating formal proofs.
To genuinely accelerate mathematicians, AI tools must integrate into this human-centric reality.
To this end, the AI co-mathematician is built to capture the enduring aspects of professional mathematical workflows, while pragmatically navigating the current strengths and weaknesses of AI reasoning.
Our design is grounded in both historical accounts of mathematical practice and our experience observing and improving previous systems for mathematical discovery, especially those developed at Google.
We relied on seven core principles to guide this design:

\begin{description}
    
\item[Embrace mathematics beyond proofs:] Genuine mathematical discovery involves a complex mixture of activities beyond theorem proving, such as iteratively refining research questions, combing the literature, brainstorming, and running extensive calculations and numerical simulations to build intuition, to name a few.
\citet{putnam1975} argued that reducing mathematics to isolated logical formalisms ignored the deeply quasi-empirical reality of the discipline, and our design deliberately reflects and supports this holistic, multi-modal reality, rather than indexing solely on final theorem proving.

\item[Support iterative refinement of intent:] In many domains, users know precisely what they want from the outset.
In mathematics, however, the process of discovery is heavily dependent on continuous iteration and refinement.
As \citet{polya1954} demonstrated in his work on plausible inference, formulating the right question often requires extensive heuristic guesswork just to get started.
As argued by Georg Cantor, ``In mathematics, the art of proposing a question must be held of higher value than solving it'' \citep[p.~27]{dauben1979}.
Experience with systems like AlphaEvolve corroborates this: mathematicians spend considerable effort actively iterating on exactly what problem to focus on---testing with smaller runs or related problems before scaling up \citep{romeraparedes2024, novikov2025}.
We designed a flexible interface that supports  multiple exploration paths, so users can fluidly refine definitions, questions, and conjectures.

\item[Produce native mathematical artifacts:] Research mathematicians already have standardized ways of working.
Rather than outputting transient chat logs or prematurely polished manuscripts, the AI co-mathematician centers its process around a living ``working paper.''
In a traditional workflow, drafting and commenting help a mathematician build a rigorous mental model of the project---knowing exactly which parts are solid and which require further attention.
Because autonomous reasoning systems can perform much of this exploratory work asynchronously out of view, the user risks losing this organic context.
To help the user reconstruct their mental model, our system tracks the evolving state of the research and visualizes it through inline text and margin notes.
By explicitly providing context---such as the provenance of specific claims or the contentiousness of a drafted lemma---the system surfaces the AI's underlying process in a format completely native to the mathematical community.

\item[Enable asynchronous interaction and flexible steering:] Mathematical research is rarely a linear process. To mirror this, the system operates not as a single conversational chatbot, but as an asynchronous team.
Under the hood, a messaging system allows multiple specialized agents to work in parallel.
Because these agents run asynchronously, the system can allocate substantial computational resources to a problem without blocking the user, allowing heavy compute and continuous interactivity to coexist.

Rather than being locked out while the system ``thinks,'' the user can communicate directly with a central project coordinator agent at any time to intervene, bypass current constraints, and steer the ongoing research.
This collaboration is also bidirectional: if the agents stall, the project coordinator agent transparently flags the roadblock and requests human assistance to help unblock the system.

\item[Manage cognitive load via progressive disclosure:] Complex research becomes unwieldy when the user and the AI agents collaboratively explore multiple parallel avenues, backtrack from dead ends, and reuse intermediate results.
A long, unstructured chat quickly becomes unusable if high-level strategy (e.g., ``Try a fixed-point approach'') is mixed with low-level execution logs of multiple background agents (e.g., ``We must verify measurability on line 40'').
To control how much information is revealed to the mathematician, our collaborative interface employs progressive disclosure, separating high-level intent from low-level execution by mirroring the agent hierarchy itself.
By default, the user interacts primarily with the project coordinator agent, which filters out the low-level execution chatter of the specialized agents' sandboxes.
However, the system always offers the user the ability to drill down into the granular details of any parallel agent's activity on demand.

\item[Track, manage, and communicate uncertainty:] Mathematical discovery demands high standards of rigor, where a single flawed lemma or hallucinated reference can undermine an entire paper.
Because the reasoning capabilities of base models are prone to unpredictable shortcomings, their zero-shot outputs introduce a baseline of uncertainty into a domain that demands precision.
Rather than hiding this friction, the system's architecture is designed around the life cycle of this uncertainty.
It \emph{tracks} uncertainty by maintaining a detailed version history, used primarily by the agents (and also available to users) to monitor how claims evolve or are called into question; it \emph{manages} uncertainty by trading compute for validation (e.g., through continuous reviews, numerical simulations, and systematic citation checking); and it \emph{communicates} friction by recognizing when the reviewing process stalls on specific parts of an argument.
By using inline highlighting and margin notes to draw the user's attention to these stalled sections, the system highlights areas of the working paper that warrant closer human scrutiny.
By treating uncertainty as a core variable to be orchestrated rather than a simple error state, robust mathematical output can be synthesized from inherently unpredictable models.

\item[Preserve the history of failed explorations:] In mathematical research, knowing what does not work is often as important as knowing what does.
While the agents are configured to navigate around roadblocks, the system accepts that specific goals or parallel workstreams may conclude in failure.
Rather than silently restarting or scrubbing the history of these unsuccessful attempts, the architecture treats dead ends as first-class, permanent outcomes.
Preserving this ``negative space'' of the project's history is crucial for tackling difficult mathematical problems.
By maintaining a durable record of failed goals and exhausted strategies, the system provides context for the user and the AI to collaboratively formulate new, more informed workstreams off the back of those very failures, reflecting the view that refutations are fundamental to mathematical progress \citep{lakatos1976}.

\end{description}

\section{AI co-mathematician in practice: a walkthrough}

To demonstrate how our design principles manifest in practice, we walk through a representative session with the co-mathematician prototype.

In this scenario, a mathematician explores an open problem in computational geometry: proving upper bounds on the area of a ``sofa'' that can move around both left- and right-handed right-angle corners.
This is a variant of a problem first introduced by \citet{moser1966moving}, with the lower bound explored in recent work on applying AlphaEvolve to mathematical optimization problems \citep{georgiev2025}.

\begin{figure}[t]
\centering
 \includegraphics[width=\columnwidth]{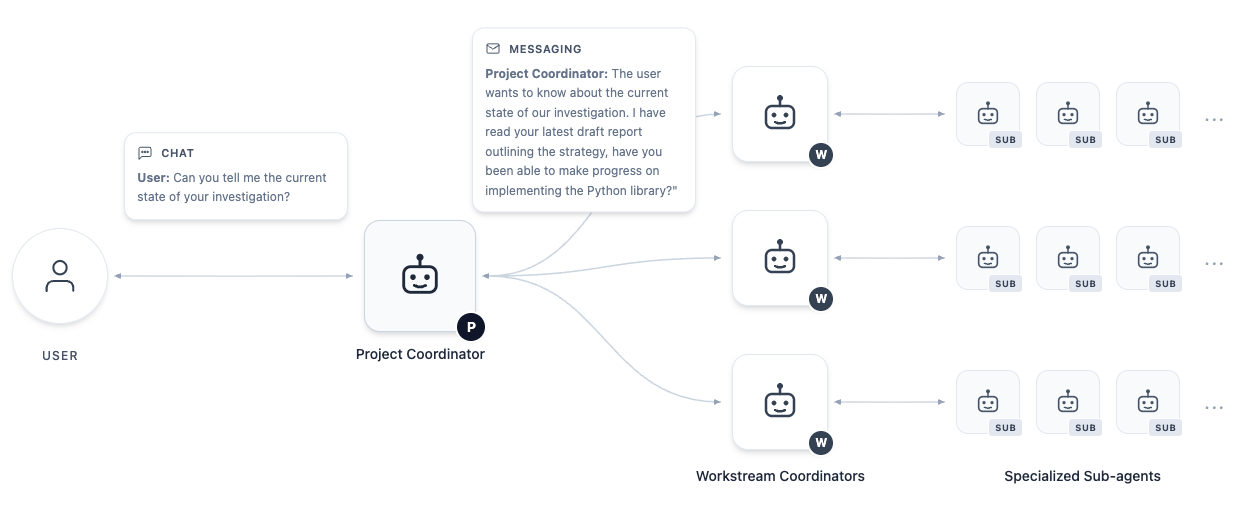}
\caption{A simplified diagram of the organization of agents in a typical AI co-mathematician workspace. Arrows denote standard communication pathways which are utilized to collect information for, and distribute instructions from, the user.}
\label{fig:org_chart}
\end{figure}

Rather than functioning as a standard conversational chatbot, the system provides an associated ``workspace'' where all information for the project is stored.
This workspace is operated on by a hierarchy of agents, which write their work to a shared file system and communicate via an internal messaging system.
The user provides input and discusses their thoughts with the top-level project coordinator agent, which delegates work to other agents further down the chain, and so on.
Much like in human organizations, we observed it to be important for the functioning of the group to have lines of communication and clear escalation pathways available to the agents, to use when they run into issues.
See \cref{fig:org_chart} for an approximate diagram of the agent organization.

\subsection{Initial Exploration}

\begin{figure}[t]
\centering
 \includegraphics[width=0.7\columnwidth]{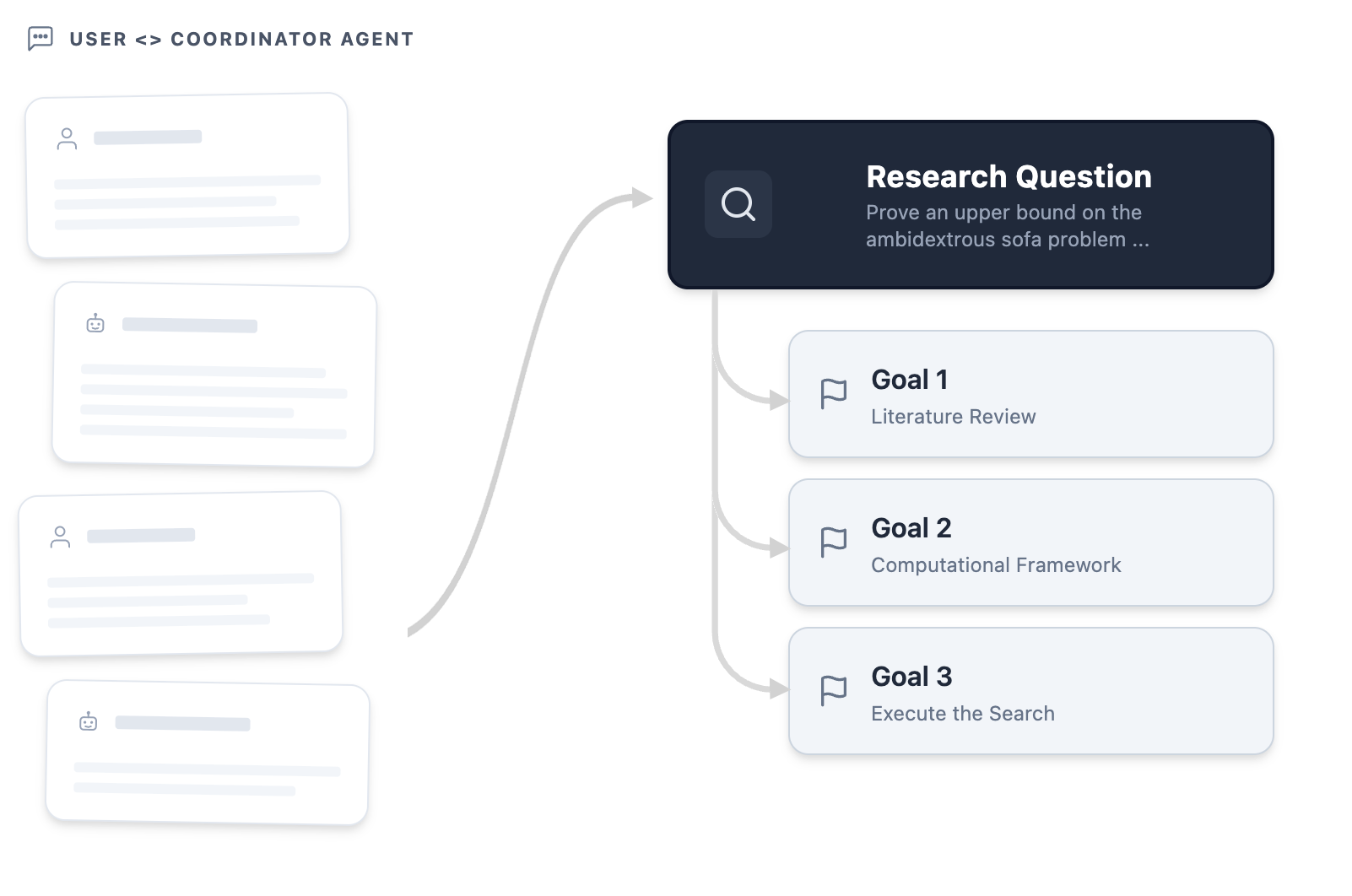}
\caption{In the initial exploration phase, a conversation between the user and the project coordinator results in an approved research question and project goals.}
\label{fig:initial_conversation}
\end{figure}

A standard zero-shot LLM requires the user to front-load a perfect prompt.
Reflecting our principle to support the iterative refinement of intent, the AI co-mathematician begins with an interactive on-boarding phase designed to help the user formalize their intent.

In this example, the user initiates the session by uploading a recent paper \citep{georgiev2025} and some context into the chat interface: \user{``I'd like to set up a project to see if we can prove some upper bounds on the variants of the moving sofa problem discussed in this paper.''}

The top-level project coordinator agent does not immediately start working on a solution.
Instead, it opens a dialogue and acts as a sounding board, replying in the chat: \user{``From the text, Baek has already proved that Gerver's lower bound is sharp for the classic sofa problem. However, the upper bounds for the other two variants are still open: \ldots{} Would you like to focus on one in particular, or both? And are you aiming to prove that a specific lower bound is sharp, or simply establish any new, rigorous upper bound?''}

Once the user clarifies, the project coordinator agent formalizes its understanding of the project's central question and high-level goals and presents them to the user.
After reading these proposals in detail, the user requests changes to the wording of the goals, and the project coordinator updates them.
At this point the user formally approves the goals, allowing the project coordinator to move on to the next phase of the project (\cref{fig:initial_conversation}).
By treating the initial interaction as a dialogue, the system ensures that downstream resources are directed toward the mathematician's actual, refined intent.

\subsection{Branching the Research}

\begin{figure}[t]
\centering
 \includegraphics[width=0.5\columnwidth]{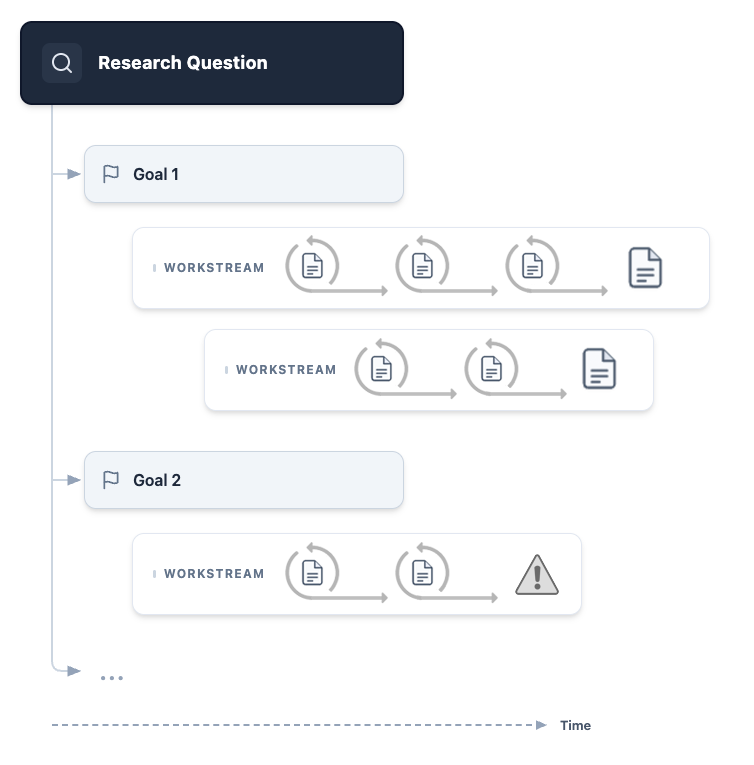}
\caption{Once the research question and goals are defined, the project coordinator schedules workstreams to make progress towards the goals. Each goal may have multiple workstreams attached to it, and additional workstreams can be added at any point during the investigation, for example in response to a specific user request. Each workstream aims to produce a fully-reviewed report complete with attachments and external references, and additionally supplies incremental reports which the user can read to judge partial progress whilst work is ongoing. Workstreams may fail to complete their work for a number of reasons; in this case, a prominent warning is displayed to the user alongside the incremental reports.}
\label{fig:multi_workstream}
\end{figure}

\begin{figure}[t]
\centering
 \includegraphics[width=\columnwidth]{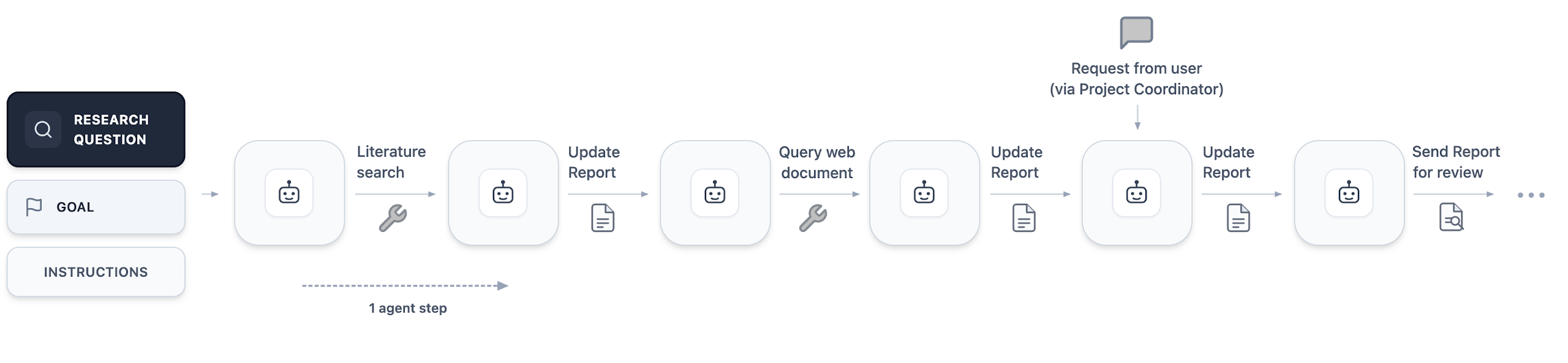}
\caption{A single workstream consists of a sequence of actions taken by a workstream coordinator agent, which may result in updates to the project state and/or user interface. In this simplified example trajectory, the workstream coordinator performs a literature search then a specific web query, each time updating the report afterwards with its findings. Following this, a message is received from the project coordinator, communicating a request from the user, which results in a further report update. The workstream coordinator then sends the latest report for review. If this passes successfully, the coordinator is able to mark its work as ``complete''.}
\label{fig:workstream}
\end{figure}

With the goals defined, the system manages the user's cognitive load via progressive disclosure: the project coordinator agent delegates work to parallel workstreams, each associated with one of the pre-approved goals, and managed by its own workstream coordinator agent (\cref{fig:multi_workstream}). As input, the workstream coordinator receives the statements of the approved research question and the selected Goal, along with any specific instructions which the project coordinator chooses to provide. 
Each coordinator performs a linear sequences of actions, which may include delegating to specialized sub-agents (\cref{fig:workstream}).
The diverse nature of these workstreams highlights the system's capacity to embrace mathematics beyond proofs.
Currently, our specialized sub-agents are based on standard LLM calls (including Gemini Deep Think) and do not make use of advanced research systems such as AlphaEvolve, AlphaProof or Aletheia, but we illustrate the points in the process where they would naturally add value.

Here we describe a scenario where the project coordinator agent starts one workstream for each goal, but it is also possible to have multiple workstreams defined per goal, or for goals to remain unpopulated with workstreams until work they depend on has been completed.

\textbf{Goal 1: Literature Review:} A workstream coordinator agent delegates to a specialized literature review sub-agent.
Utilizing our computationally intensive search tool, it discovers the key papers containing the machinery used for previous upper bounds in the sofa problem.
With pointers to important references, it queries them directly using specific web and literature access tools, to identify the exact statements and proofs of key results.

\textbf{Goal 2: Computational Framework:} Another workstream coordinator agent tackles the task of designing the required computational framework.
Based on strategies observed in prior literature, the workstream coordinator agent first designs the high-level logic for the computational framework, using a sub-agent creation tool to dispatch Gemini Deep Think to prove that this could give a rigorous upper bound as required.
Once this is proven, it uses another sub-agent creation tool to create a coding agent, instructing it to implement the required Python library with associated tests and demonstration cases.
As each of these agents gives status updates, the workstream coordinator agent asynchronously updates the user-facing workstream report with details of the proofs and provides links to the key code files.

All proofs in the existing system are purely informal, but the possibility of adding a formal prover such as AlphaProof would allow for increased confidence in the correctness of proofs, for those that are able to be stated and proved with existing formal mathematics libraries.
Alternatively, using a stronger informal proving system such as Aletheia would add considerable power to the system, at an associated computational cost.

\textbf{Goal 3: Execute the Search:} The final branch-and-bound search is executed in a further workstream, created by the project coordinator agent after the computational framework workstream has successfully completed.
The shared file system allows this workstream coordinator agent to import the library developed in the previous workstream, and it now uses a parallel code execution tool to scale up the search beyond a single script.
Parallel code execution requests are run on separate cloud machines, avoiding the burden for the user to set up or access a local cluster.
The results of the search are collated into a single file which is attached to the workstream report, with summary information and conclusions written up into the main text.

In the current prototype, such computational searches are executed as a combination of standard agent primitives, but the addition of evolutionary search algorithms such as AlphaEvolve would provide a supercharged engine for such optimization and search routines.

\subsection{Interactive Steering and Hard Constraints}

When tasked with solving difficult research problems, standard AI agents often find invalid shortcuts, hallucinate lemmas, hand-wave details, and claim success prematurely.
One of the defining technical features of the AI co-mathematician is the application of hard programmatic constraints to prevent these failure modes, combined with active human steering by the user.

During the computational exploration, if the search space explodes, the naive backtracking approach initially proposed may fail.
In this situation, the coding sub-agent is bound by strict rules: it cannot mark code as finished until its tests pass and a reviewer agent accepts the validity of the code and golden values in the tests.
Because the reviewer agent repeatedly rejects the failing code, the workstream coordinator agent is blocked.
Rather than silently restarting, the system preserves this failed exploration as a durable record in the shared file system, which the project coordinator reads.

Forced to manage and disclose its uncertainty, the project coordinator agent surfaces an alert to the user and explicitly asks for help in the chat: \user{``Our initial implementation of the search is not efficient enough to find the result that we need, and there aren't many other examples in the literature. Do you have a mathematical intuition for a better pruning strategy we can apply? More details on the current approach can be found in the following document.''}

Crucially, the user is not locked out while agents work.
Via the chat interface, the user reads the project coordinator's update and actively steers the project, sending a message to suggest a topological pruning heuristic, and directing the coordinator to create additional workstreams to investigate new bounding strategies while the current work is ongoing.

\subsection{The Final Output}

After the user steers the system past the computational bottleneck, the workstreams conclude their specific goals.
However, the final output is not a transient chat message.
Each workstream coordinator agent is responsible for producing, as its main output, a compiled and reviewed LaTeX write-up, which is displayed prominently to the user when they open the workstream.

In order to produce natural mathematical artifacts, we require the coordinators to meet specific criteria in these write-ups:

\begin{description}
\item[Exposition:] The draft must include an explanation of the research process that led to the outcome, not just the final result.

\item[Margin Annotations:] The document utilizes margin notes to provide additional information, explicitly linking claims back to the workspace.
A note might read: \user{[Pruning heuristic derived from user suggestion; baseline bound of 2.2195 sourced from paper at arxiv.org/abs/...].}

\item[Internal Linking:] As well as external literature, references are made to internal documents created by the agents, giving the user direct entry points to audit the shared filesystem.

\item[Review Process:] Before a report can be marked as finalized and the workstream completed, the report must be submitted to a paper review process, where it is scrutinized for content and style by several AI reviewer agents, all of whom have tools and capabilities to cross-check references and code outputs as well as logical correctness.
Reviewer agents persist between review rounds, creating an iterative process over which the report is refined and improved, which only concludes once \textit{all} reviewers have formally approved the report.
To avoid infinite loops, if the workstream coordinator agent is unable to pass the review process it can escalate this issue to the project coordinator agent.
In this case, the workstream is marked as unfinished, with an escalation message clearly surfaced to the user, allowing the user to immediately understand that this report may contain unresolved issues.
\end{description}

\section{The Evaluation of Interactive Mathematical Agents}
\label{sec:evaluation_paradigm}

Historically, AI-for-mathematics progress has been measured against static problem-solving benchmarks.
Today, the field relies heavily on such benchmarks, including IMO ProofBench \citep{luong2025imobench}, FrontierMath \citep{glazer2024}, and PutnamBench \citep{tsou2024putnambench}, and their predecessors MATH \citep{hendrycks2021math} and GSM8K \citep{cobbe2021gsm8k} were highly influential in measuring early LLM progress.

However, with frontier systems now demonstrably performing at or above the level of expert humans at \textit{raw problem-solving ability} \citep[e.g.,][]{luong2025advanced, abouzaid2026proof}, we believe the incremental value of improving this ability is reducing, and that progress in AI-assisted mathematics is now at least partially bottlenecked on other capabilities that reflect the broader workflows of professional mathematicians.
Similar evaluation shifts are emerging across other areas of science, measuring real-world success directly by running wet lab experiments \citep{gottweis2025towards} and solving open problems \citep{woodruff2026accelerating}.

This implies, firstly, that AI systems should be measured against a wider array of tasks. 
Benchmarks such as DeepSearchQA \citep{gupta2026} for general purpose fact-finding, and Hard2Verify \citep{pandit2025} for mistake-finding in competition-mathematics proofs, measure capabilities with exact analogues in the mathematical research process.
But to our knowledge no similar benchmarks have been developed for the setting of professional research mathematics.
Clear measurements of these auxiliary capabilities (and improvements against them!) will become increasingly important as AI systems improve, to reduce the severity of the ``jagged frontier'' \citep{dellacqua2026navigating} and the associated impedance mismatch between human-driven and AI-driven mathematics.

And yet more importantly, we believe that AI-for-mathematics systems should be considered as human-in-the-loop by default, and advancements and achievements measured accordingly.
This echoes our design principle that the foremost purpose of our system is to assist mathematicians with their work, not to achieve success independently of them.
With this framing in mind, we first present results which have arisen from mathematicians directly using the AI co-mathematician (\cref{sec:early_results}), followed by measurements of the AI co-mathematician's static problem-solving performance (\cref{sec:benchmarks}).

\section{Early Results with Mathematicians}
\label{sec:early_results}

As part of our efforts to make interactive tools more widely available, we have given access to the AI co-mathematician to a small number of professional mathematicians to use for their own research.
The wide range of use cases that early users have explored with the AI co-mathematician reflects its broad applicability and integration into standard research workflows.
The AI co-mathematician has served as a functional aid in instances such as navigating disparate literatures, performing numerical experimentation, and obtaining proofs across several domains in mathematics.
In the remainder of this section, we showcase several use cases illustrating the system's current utility, obtained by recent users in our limited release.
It should be stated that whilst many users have had successful interactions with the AI co-mathematician leading to novel discoveries, there has been a range of feedback and satisfaction with the tool, and other users have found it less effective for their work.
We discuss some of the challenges in section \ref{sec:limitations}, and we hope to use the experiences and feedback from these mathematicians to inform future benchmarking and development directions.

Notably, all results showcased here were produced independently by mathematicians using the tool directly, without any supervision or intervention from GDM researchers.

\subsection{Case Study: A Kourovka Problem}

An early user, M. Lackenby, used AI co-mathematician to investigate several problems in topology and group theory.
His work led to a resolution of an open question (Problem 21.10 from the Kourovka Notebook).\footnote{The problem asks whether every finite group admits a so-called ``just finite presentation,'' i.e., a finite presentation where removing any relation results in an infinite group. The answer turned out to be affirmative.}
Lackenby's process in this case illustrates the value of a ``mathematician-in-the-loop'' style of interaction.

Lackenby began by simply typing in the problem statement; the system confirmed the meaning and then created two independent workstreams, one attempting to prove the result, the other to disprove it.
The first result to come back was a ``proof'' that the system itself marked as incorrect---a workstream had written up an argument, but then a reviewer agent had found a flaw.
However, when looking at the paper, Lackenby realized that despite this gap, it contained what he called a ``really, really clever proof strategy.''
Furthermore, when he read the reviewer's critique of the proof, he realized, ``Hang on a second, I know how to fill that gap.''

He pointed out a way to correct the argument, and the system was able to write up a complete and correct proof.
Finally, Lackenby downloaded the document and applied his own revisions---generalizing the result and adding examples---before uploading it again and asking the project coordinator to create a workstream for a final review.
It did so, spotting two minor issues in the paper that he fixed for the final copy.

A lesson from this case is that the back-and-forth between AI and mathematician was crucial to the resolution of the problem.
To Lackenby, this suggests ``the system works best when the user is familiar with the area.''
While one might see this as a limitation, he adds, ``What's the point in getting it to solve a problem that I have no idea about?''

\paragraph{Agent Actions}

Behind the scenes, the coordinator agent for the workstream:
\begin{itemize}
     \item Created a coding sub-agent to perform a computational search for non-just-finite presentations, which succeeded in finding two examples.
     \item Analyzed these and performed literature searches to identify relevant theory to their constructions, leading to a general method of producing non-just-finite presentations.
     \item Noted that this did not negatively resolve the conjecture (as groups may have multiple presentations, and finding one non-just-finite presentation does not imply that none of them are just-finite), but decided to conclude its workstream here with a sharper conjecture based on this argument, and sent the report for review.
     \item Opened a back-and-forth discussion with reviewer agents as part of the review process.
     During this, one of the reviewer agents spotted a method that could be used to prove the conjecture in the positive direction.
     \item Agreed with the reviewer and pivoted the workstream to writing up this method.
     After further reviews, this led to the draft which Lackenby augmented to complete the proof.
\end{itemize}

\subsection{Case Study: Stirling Coefficients}

Another early user, G. B\'erczi, approached the AI co-mathematician by initially tackling a problem
concerning the behavior of Stirling coefficients for symmetric power representations.
Specifically, the conjectures hypothesized that in a certain binomial expansion, the coefficients are not only strictly positive but also form a log-concave sequence.

To initiate the project, B\'erczi prepared a brief note introducing the topic, the background of the
conjectures, and known methods.
This primer also included suggested directions derived from his prior experiments with other systems, such as AlphaEvolve.
While AlphaEvolve had failed to resolve the higher-index cases, it had hinted at a possible direction to recover an inductive formula for the coefficients.
B\'erczi included this in his document, which he uploaded as part of the initial project formulation discussion.
This style of providing deep, rich background on the problem, which B\'erczi referred to as ``structured posing of questions,'' was one that he had found effective with other AI systems.

In response to B\'erczi's question, the AI co-mathematician established proofs (currently in detailed human review) for two of the conjectures in separate workstreams.
It additionally provided detailed computational evidence for its claims as well as for investigations into the unproven conjectures.
B\'erczi noted that the design of the system helped in various ways.
Not only was it easy to follow the ``green ticks'' checking off tasks, but more importantly, a marginal comment in the finished document alerted him to a key insight, which he was able to ask about in the chat interface.
Nonetheless, he pointed out that collaborating with AI systems requires some skill, saying ``It's not trivial how to use this now'' and suggested that ``It will make a big difference between mathematicians, how they use these models.''

\paragraph{Agent Actions}

Two key workstreams provided claimed proofs of the conjectures.
To illustrate the workflow, in the first workstream, the coordinator agent:
\begin{itemize}
     \item Created a coding sub-agent to perform an initial computational enumeration of expansions, to an extent that was reasonable to run in a few minutes.
     \item Observed from the results that the conjecture as stated was false for $n = 1, 2$, but appeared to hold for larger $n$, and further that its original proof strategy was invalid.
     \item Used a Gemini Deep Think sub-agent to propose a new proof strategy for the updated conjecture.
     It did so successfully, convincing the workstream coordinator agent and subsequently the reviewer agent.
\end{itemize}

\subsection{Case Study: A Lemma in Hamiltonian Systems}

A third early user of the AI co-mathematician, S. Rezchikov, posed a recent technical subproblem that he had come across in his research, on the topic of the existence of perturbations of a specific class of Hamiltonian diffeomorphism with certain convenient properties.

Using the system, Rezchikov discussed the problem with the project coordinator agent and provided supplementary papers which contained results relevant to the problem statement, until agreement was reached on the precise definition of the task, and a workstream created in order to attempt to solve it.
The write-up produced by the workstream included a key lemma with an elegant proof, that withstood careful checking, and essentially resolved the question posed by Rezchikov.
Again, Rezchikov noted that other AI systems had failed to prove the result given the same prompt, though without multiple sampling attempts and controlled conditions it is hard to draw conclusions from such individual experiments.

Rezchikov made two other observations which shed light on the value of the system.
One was about exploration of an approach which did not seem to work: here the added value was that the system let him reach a dead end faster than otherwise.
As he put it, ``I could have easily spent a week dreaming about what was there, but instead I just moved on.''
Separately, he noted that the quality of correct proofs seemed comparatively high: ``I would rank, aesthetically, its general style of proofs as the best one of any models I've gotten to use.''

This gives an illustration of how the AI co-mathematician can help mathematics researchers explore ideas in a manner that integrates naturally into their typical exploratory research workflows.

\paragraph{Agent Actions}

In this case, the workstream coordinator agent:
\begin{itemize}
    \item Performed a literature review using the specialized tool, into commonly used techniques and pitfalls for the problem in question, which highlighted a few key properties to be aware of in addition to those provided by Rezchikov.
    \item Made specific follow-up literature queries to understand the precise nature of these points.
    \item Passed the problem statement and the key pieces of context to Gemini Deep Think, which produced the proof, including the key lemma, which was then written up into the report.   
\end{itemize}

\section{Problem-Solving Benchmark Results}
\label{sec:benchmarks}

Having made a case for broader measurements of abilities, we acknowledge that problem-solving benchmarks are currently the best available objective measures of AI systems' performance in mathematics.
In this section, we discuss our attempts to measure the AI co-mathematician's performance using such benchmarks.

As the AI co-mathematician is designed to communicate results via long-form reports and ask questions of the user when stuck, we developed a custom mode where the system is not able to receive external input beyond that of an initial question, and returns a single final answer in the user chat panel.
The main differences implemented for this mode are:
\begin{itemize}
    \item Bypassing of the initial problem-definition conversation, with the assumption that all the required information is present in the problem statement.
    \item Hard-coding of a single project goal to ``solve the problem''.
    \item 
    \begin{minipage}[t]{\linewidth}
    The introduction of a fixed time limit, after which the project coordinator agent is required to give a final answer, if it has not already. This was set to 24 hours for internal evaluations and 48 hours for FrontierMath.
    Most problem attempts finish well within this time limit, but for hard problems without human intervention the system occasionally continues beyond this, especially in cases where it struggles to produce a clean solution.
    \end{minipage}
\end{itemize}

As a long-running agentic system, compared to base models which are typically evaluated on these tasks the AI co-mathematician uses a correspondingly larger amount of compute. However, though there are no explicit inference limits in the AI co-mathematician solution attempts, the system was developed to be efficient enough to serve to many external users. We observe that each attempt uses a broadly comparable number of model and tool calls to a long AI-assisted software engineering session, matching its primary use case as an interactive agentic tool.

\subsection{Internal Research Mathematics Evaluation}

We tested the AI co-mathematician on an internal benchmark consisting of 100 unleaked, research-level mathematics problems with code-checkable answers, sourced from professional mathematicians.

\begin{figure}[t]
\centering
 \includegraphics[width=0.85\columnwidth]{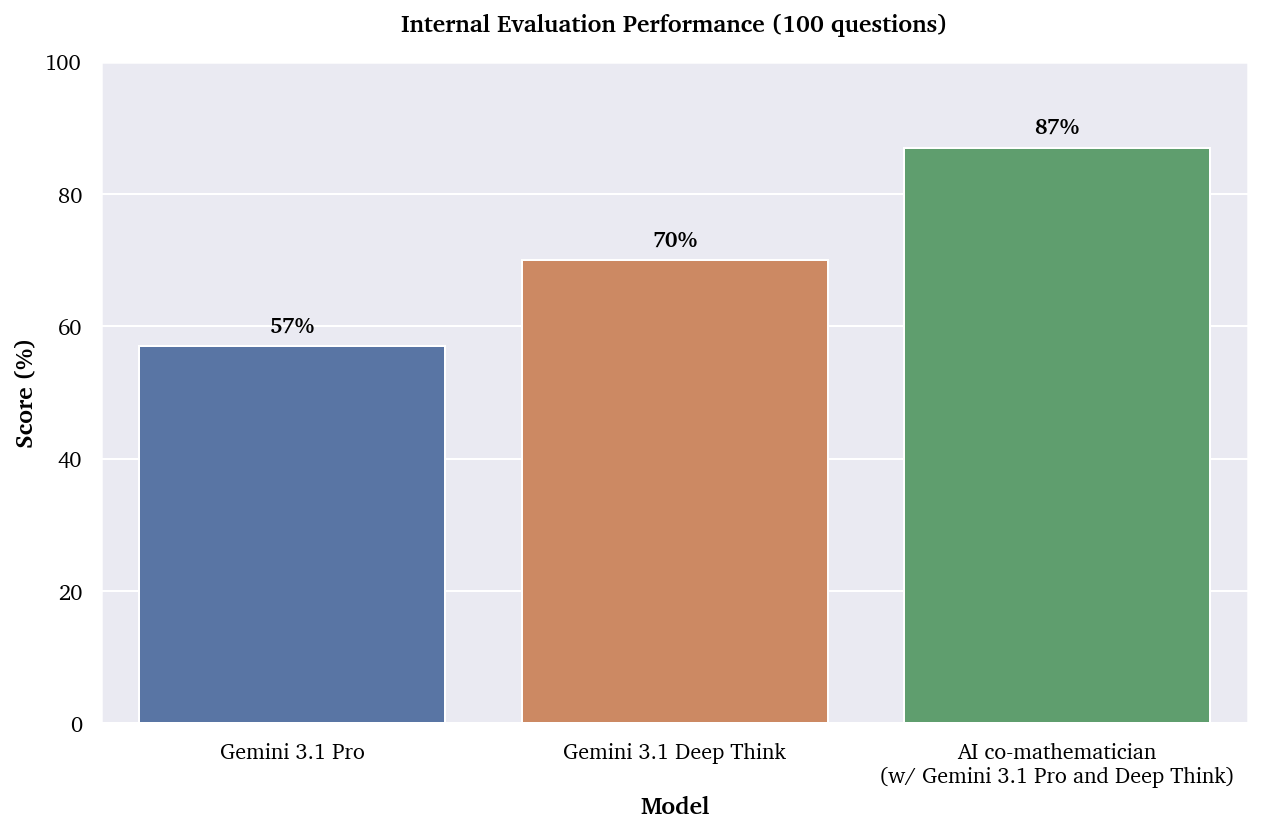}
\caption{Accuracy scores for Gemini 3.1 Pro, Gemini 3.1 Deep Think, and the AI co-mathematician (also based on Gemini 3.1), on an internal research mathematics benchmark.}
\label{fig:benchmark}
\end{figure}

On these benchmarks, the AI co-mathematician significantly outperforms a single call to the Gemini 3.1 Pro model, and a single call to the Gemini Deep Think system (\cref{fig:benchmark}). As discussed, the AI co-mathematician also uses a larger amount of compute: indeed, many of the internal agents underlying the AI co-mathematician are built on Gemini 3.1 Pro, and the prover agent can utilize Gemini Deep Think.

Analyzing specific examples, we see the benefits of various features of the AI co-mathematician.

In one question relating to geometric tilings, we see that the AI co-mathematician successfully reduces the core of the question to a Boolean satisfiability (SAT) problem, which it solves using the \texttt{PySAT} library.
Several non-trivial adaptations were required to the core problem to make this framing work---this form of solution clearly favors a system which has a persistent file system and agentic code-writing abilities.
Whilst the evaluation harness for other models allows code to be executed, developing and testing more complex libraries is difficult for them to achieve.
Both Gemini 3.1 Pro and Gemini Deep Think attempt a purely theoretical solution to this problem, which in this case is a far harder task.

In a problem on representation theory, the AI co-mathematician uses its literature search tools to correctly retrieve exact theorem statements from relevant papers, and apply them to solve the task.
The other models attempt to apply more general theorems; though these are relevant to the subject area, without literature access to check precise statements they aren't able to correctly match the theorem conditions to the assumptions present in the problem.

In a combinatorics problem with both a theoretical and computational aspect, we see the AI co-mathematician cleanly separates its work into multiple workstreams, refining the theoretical foundation for the solution separately from the development of code.
In the theoretical workstream, multiple logical errors and inconsistencies (similar to those made by the other models) were pointed out by the reviewers running spot-checks, and were eventually corrected for the final workstream report, allowing the project coordinator to neatly assemble the pieces to give a correct solution.

\subsection{FrontierMath}

A natural comparison is the external FrontierMath benchmark \citep{glazer2024}.
Epoch AI tested the AI co-mathematician on Tier 4 of the benchmark, in the final-answer mode described above. In the words of Epoch AI, ``This extreme tier consists of 50 problems crafted as short-term research projects by professors and postdocs. They are designed to surpass Tier 3 in difficulty, with some potentially remaining unsolved by AI for decades''. Frontier models and systems from all major providers are regularly tested on the FrontierMath benchmark.\footnote{Latest results and leaderboards can be viewed at \url{https://epoch.ai/frontiermath/tiers-1-4?tier=Tier+4}.}

The evaluation was conducted blind, with Epoch AI having direct access to the AI co-mathematician UI and entering the problems and retrieving the answers themselves, without the system developers seeing the problems or observing the state of the project workspaces during the evaluation.

The AI co-mathematician achieved a new high score on this benchmark, correctly solving 23 problems out of 48, once the two public sample problems are excluded, an accuracy of 48\%. Notably, this is a significant increase over the Gemini 3.1 Pro base model which achieved 19\% accuracy and which the AI co-mathematician is based on, showing our parallel investigation branches, enforced review cycles and tool implementations lead to improved outcomes on problem-solving tasks. The problems solved correctly included three problems which had not been solved by any previously evaluated system, though the AI co-mathematician also failed to solve two problems which had been previously solved by at least one system. 

An important difference versus prior evaluations is that FrontierMath evaluations are generally executed with a standard agentic harness developed by Epoch AI.\footnote{A notable exception: Gemini 2.5 Deep Think was also evaluated via the web UI without an additional harness \citep{epoch2025evaluatingdeepthink}.} This offers access to a Python interpreter but also places a hard limit on the number of tokens used in the agent trajectory (to the extent that these are accounted for in the associated model API). In our setup however, we only use our own tool implementations and place no limit on the number of model calls or tokens generated. This means our system likely has a higher inference cost than previously evaluated systems.

\section{Challenges and Limitations}
\label{sec:limitations}

While the AI co-mathematician demonstrates the potential of interactive AI workflows, we encountered several difficulties in making such a system genuinely useful.

\begin{description}

    \item [Reviewer-Pleasing Bias (False Consensus):] The iterative review process is a dynamical system, evolving the working documents.
    When an agent produces a flawed argument that it cannot genuinely fix, the strict constraint of satisfying the reviewer agents can sometimes cause this system to converge to an argument that remains flawed, but where the errors can no longer be detected by the reviewer agent.
    Such arguments can also be tricky for humans to tease apart, which has been highlighted as a limitation of current AI systems \citep{kontorovich2025shapemathcome}.
    Similar pathologies in prover--verifier dynamics have been noted in the literature \citep{dang2026escapingcognitivewellefficient}.
    Although this behavior is relatively rare, it represents a violation of our core principle of explicitly acknowledging uncertainty. 
    
    \item[Intractable Disagreements (Non-Termination):] Conversely, when the iterative review process fails to reach consensus, it can fail to terminate entirely.
    Under these dynamics, the iterative review process becomes locked in an endless cycle of revisions and rejections.
    Over successive autonomous iterations, this loop often degrades into increasingly hallucinated reasoning---a phenomenon colloquially known as a ``death spiral.''
    While we have implemented various mechanisms to mitigate these effects, the core issue of language models frequently disagreeing with each other remains.
    Early users have noted this behavior, learning to recognize when a workstream has entered such a state and appropriately down-weighting their trust in its output.

    \item[System Autonomy Requires Ceding Control:] Mathematical research is inherently exploratory; predefined task planning is often impossible because discovering the correct sequence of steps is synonymous with solving the problem itself.
    Consequently, there is always a significant risk that the model encounters an unplanned-for difficulty, which is amplified in the AI co-mathematician where we allow the system to work autonomously for many hours without requiring intermediate human input.
    It is clear from our experiences that current models' judgment on what to do in such cases is far behind human capabilities and expectations, and it is a challenging balancing act to allow long-running autonomous capabilities while retaining user controllability.

    \item[Semantic Meaning of Representations:] Our team recognized early on that mathematicians tend to associate a well-typeset document with a corresponding level of rigor in the content, an intuition that is belied by LLMs, which excel at generating flawless LaTeX formatting while frequently struggling with rigorous logical verification.
    In the AI co-mathematician, we attempt to reduce this by highlighting the output as a ``working document'' with associated marginalia, but it seems that new interfaces could be developed to more intuitively represent the characteristics of such documents.
    Recent AI-enabled research in HCI \citep[e.g.,][]{pair2025lumi} could provide inspiration for the path forward.
    
\end{description}

Further, the introduction of highly capable, agentic systems into the mathematical ecosystem carries some risk.
Moving forward, the community must navigate several systemic challenges:

\begin{description}

    \item[Maintaining Signal-to-Noise in the Literature:] As AI systems become highly proficient at typesetting LaTeX and synthesizing literature, the community will need to manage a potential influx of automated output.
    If these tools are frequently used as autonomous generators rather than human-steered collaborators, the field would see a rise in plausible but shallow, incremental, or subtly flawed papers.
    This shift could increase systemic ``semantic noise,'' requiring researchers to develop new heuristics to efficiently identify genuinely novel mathematical discoveries amidst a higher baseline volume of drafts.
    Formal methods and auto-formalization could plausibly assist with checking correctness and highlighting flaws in papers, but will not substitute for the process of community members understanding and absorbing the implications of newly published works.
    
    \item[Adapting the Peer Review Ecosystem:] The mathematical peer review process relies on deep, intensive human verification.
    Agentic AI introduces a challenging new scaling dynamic: a system can generate a 20-page proof attempt in minutes, but a human expert may take days to verify it.
    If AI-assisted drafting becomes ubiquitous without built-in, auditable paper trails, it could place a heavy and disproportionate burden on the volunteer peer review system.
    The AI co-mathematician's margin annotations are a small initial step toward increasing the auditability of results, but broader community standards will be needed.
    Furthermore, although our system features specialized reviewer agents, substituting AI for human referees presents its own risks.    While automated reviewers excel at local logical checks, catching algebraic slips, or surfacing missing citations, they currently lack the holistic intuition required to judge a paper's elegance, depth, or true mathematical significance.
    Over-relying on AI for peer review risks reducing mathematical evaluation to mechanical verification, potentially sidelining the qualitative human judgment that guides the field.

\end{description}

\section{Conclusion}

The AI community has recently passed multiple technical milestones, demonstrating human-level and superhuman performance across a variety of mathematical benchmarks.
If we want to genuinely accelerate scientific discovery, however, solving static, well-defined problems is only part of the solution.
The frontier of mathematical research is not a linear conversation or a series of disconnected puzzles; it is a messy, overlapping, highly iterative process defined by unproven intuition, branching hypotheses, and complex human collaboration.

The AI co-mathematician is designed to meet researchers where they are.
Rather than reducing AI to an isolated oracle or a simple verification pipeline, the system acts as a holistic workspace.
By managing the life cycle of uncertainty, mirroring human workflows through hierarchical delegation, and grounding its output in native mathematical artifacts, it elevates powerful base models into natural collaborators.
Through hard programmatic constraints and the continuous maintenance of a living ``working paper,'' the system ensures that the full research journey---the failed tests, the literature synthesized, and the continuous refinement of an idea---is captured, audited, and explicitly communicated to the user.
As demonstrated by early users resolving open problems and finding novel proofs, this bidirectional exchange allows humans to effectively steer AI past difficult bottlenecks.

Realizing the full potential of this paradigm, however, requires a shift in how the AI community measures success.
While these benchmarks are excellent at evaluating a model's ability to generate a final answer to a curated problem, they do not capture the full scope of frontier research.
They are not designed to measure a system's ability to interactively prune a hypothesis tree, synthesize niche literature, or appropriately halt and disclose uncertainty when scaling fails.

If we want to build systems that act as true co-mathematicians, we must develop complementary evaluation frameworks that measure collaborative efficacy, stateful exploration, and the rigorous management of uncertainty.
The next revolution in AI-assisted mathematics will not just be defined by which model can synthesize the right answer the fastest, but by which system can most effectively help human researchers navigate the unknown.

\section*{Acknowledgments}

We would like to thank:

Edward Lockhart, Allison Woodruff, Juanita Bawagan, Uchechi Okereke, Thang Luong for reviewing this report.

Anna Trostanetski, Andrey Petrov, Matin Akhlaghinia, Victoria Johnston, Nick Dietrich for their help in improving the deployment setup and reliability of the prototype system.

Mariana Felix, Francesca Pietra for advising on externalization and partnerships.

Uchechi Okereke, Gemma Gibbs for legal counsel. 

Adriana Lara, Armin Senoner, Danielle Breen, Duncan Smith, Juanita Bawagan for advising on communications strategy and naming.

Henryk Michalewski for maintaining the internal dataset of research mathematics problems.

Greg Burnham (Epoch AI) for coordinating the FrontierMath evaluation.

Ellen Jiang for UI advice and improvements.

Victoria Johnston, Doug Fritz, Felix Riedel for frontend code reviews. 

Sébastien Racaniere, Romu Elie for communicating with early testers and members of the mathematical community and feeding back insights and requests from them.

Richard Bamler, Johannes Bausch, Mehdi Bennani, Gergely Bérczi, David Berghaus, Otis Chodosh, Bennett Chow, Maria Chudnovsky, Romu Elie, Sergey Galkin, Javier Gómez-Serrano, Matt Harvey, Amaury Hayat, Marcus Hutter, Ray Jiang, Ayush Khaitan, Alex Kontorovich, Robin Kothari, Marc Lackenby, Tor Lattimore, Igor Makhlin, Johan Martens, Alex Matthews, Stanislav Nikolov, Georg Ostrovski, Stan Palasek, Sébastien Racaniere, Danylo Radchenko, Johannes Ruf, Julian Salazar, Simone Severini, Phiala Shanahan, Iain Smears, Elahe Vedadi, Adam Zsolt Wagner, for testing the system and providing feedback.

Javier Gómez-Serrano and Terence Tao for early testing and feedback on the literature search capabilities.

Nada Baessa, Bruno Vergara Biggio, Semon Rezchikov for in-depth testing, checking of outputs, detailed feedback, feature requests.

Allison Woodruff, Patrick Gage Kelley for help with interviewing early users and collating feedback.

Marc Lackenby for feedback on the mathematical performance of the system, input into the workflow and interface design, and deep mathematical collaboration.

JD Velasquez, Yunhan Xu for advising on distribution and product viewpoints.

Victor Martin, Stig Petersen, Petko Yotov, Hamish Tomlinson, Sam Blackwell for sourcing compute and assisting with model serving.

Sam Blackwell for engineering support and advice.

Stig Petersen for technical reviewing.

The IAS at Princeton for hosting a joint workshop last year, at which many of these topics were discussed.

Gemini 3.1 Pro was used at various stages in the preparation of this manuscript, for drafting portions of the main text, proofreading human-written sections, and producing and iterating on figures and diagrams.

\section*{Contributions}

Y.Z. designed the initial prototype of the system. 
Y.Z., I.V.G., D.Z., I.B., L.B., A.D. continued development of the core system and user interface and performed research on improving capabilities, with input from V.K. on research ideas.
M.W., D.K. provided input and testing throughout prototype iterations.
M.W. designed key elements of the user interface and workflow, F.V. designed the visual language and further UI elements.
T.S. developed the route for external users to access the system.
S.A. developed the system requirements for externalization.
I.V.G., Y.Z., D.Z., I.B., L.B., A.C., T.S. maintained the system for external users.
Y.Z., I.V.G., D.Z., I.B., L.B., D.M.R., H.M. prepared the system for internal and external evaluations.
D.M.R., M.W., D.Z. wrote the paper, with input from I.V.G., Y.Z., A.D., B.G., F.V.
F.V., M.W., D.Z., D.M.R. created the diagrams.
B.G. coordinated the external user community.
D.Z., I.V.G., and G.H. coordinated the team.
A.D. and D.Z. developed the overall strategy.
A.D. and P.K. supervised the research program.

\printbibliography

\end{document}